\documentclass[10pt,twocolumn,letterpaper]{article}

\usepackage[pagenumbers]{wacv} 

\usepackage{graphicx}
\usepackage{amsmath}
\usepackage{amssymb}
\usepackage{tabularx}
\usepackage{booktabs}

\usepackage[pagebackref,breaklinks,colorlinks]{hyperref}

\usepackage[capitalize]{cleveref}
\crefname{section}{Sec.}{Secs.}
\Crefname{section}{Section}{Sections}
\Crefname{table}{Table}{Tables}
\crefname{table}{Tab.}{Tabs.}
\usepackage{color} 

\def\wacvPaperID{2428} 
\def\confName{WACV}
\def\confYear{2025}

\begin{document}

\title{Real Classification by Description: Extending CLIP's Limits of Part Attributes Recognition}


\author{%
  Ethan Baron\textsuperscript{1}, Idan Tankel\textsuperscript{1}, Peter Tu\textsuperscript{2}, Guy Ben-Yosef\textsuperscript{1} \\[1ex]
  \textsuperscript{1}GE HealthCare Technology and Innovation Center \\
  \textsuperscript{2}GE Aerospace Research
}
\maketitle

\begin{abstract}
   In this study, we define and tackle zero-shot 'real' classification by description, a novel task that evaluates the ability of Vision-Language Models (VLMs) like CLIP to classify objects based solely on descriptive attributes, excluding object class names. This approach highlights the current limitations of VLMs in understanding intricate object descriptions, pushing these models beyond mere object recognition. To facilitate this exploration, we introduce a new challenge and release description data for six popular fine-grained benchmarks, which omit object names to encourage genuine zero-shot learning within the research community. 
   Additionally, we propose a method to enhance CLIP’s attribute detection capabilities through targeted training using ImageNet21k's diverse object categories, paired with rich attribute descriptions generated by Large Language Models (LLMs). Furthermore, we introduce a modified CLIP architecture that leverages multiple resolutions to improve the detection of fine-grained part attributes. 
   Through these efforts, we broaden the understanding of part-attribute recognition in CLIP, improving its performance in fine-grained classification tasks across six popular benchmarks, as well as in the PACO dataset, a widely used benchmark for object-attribute recognition.
   \footnote{Code: \url{https://github.com/ethanbar11/grounding_ge_public}}

\end{abstract}

\label{lebl:intro}
\section{Introduction}

Zero-shot classification is increasingly recognized as a key capability of Vision-Language Models (VLMs) and Large Multimodal Models, showcasing their ability to accurately classify objects or concepts not explicitly encountered during training \cite{radford2021learning, yuksekgonul2022and, menon2022visual, roth2023waffling, pratt2023does}. This capability highlights the power of generalized representation learning.  

Moving beyond traditional VLMs' reliance on direct matching of category labels to images, the emerging trend of classification by description (e.g., \cite{menon2022visual, Kaul2023}) offers a more nuanced approach. This method, gaining traction for its explanatory potential, utilizes Large Language Models (LLMs) to generate more complex attribute-based descriptions of objects' internal components and features. When integrated with CLIP's pre-training methodology, these models adeptly compute cosine similarities between rich textual descriptions and images, aiming to provide a deeper, more interpretable understanding of visual data. The efficacy of this approach is often tested across various object classification benchmarks, including benchmarks for fine-grained categorization, stressing the significant potential of zero-shot classification to harness the synergies between language and vision for more explainable and effective AI solutions.

The ability to identify objects based on their descriptions is a fundamental aspect of human cognitive capability, a skill that VLMs are striving to emulate. However, as we show below, models like CLIP, while dominant in the field, fall short in this regard. Our experiments show that they often fail to accurately identify objects solely on the basis of descriptive text, highlighting a notable gap between current AI capabilities and human-like comprehension. Similar gaps were also shown in other works on the general semantic attribute-based description of images (e.g.,~\cite{yuksekgonul2022and}, \cite{zhao2023vlchecklist}), but here we specifically focus on the semantic descriptions of objects and their parts, in free-form language.


In this study, our main research question centers on the potential of training VLMs to understand object descriptions in terms of their internal parts and visual attributes. If a model is trained on attributes (and most importantly, part attributes) coming from one set of classes, how well can it generalize the understanding of attributes to novel classes? To achieve this, we utilize LLMs to mine essential 
descriptions of objects from the diverse ImageNet21k dataset. This approach allows us to gather a wide range of descriptive texts, not limited to clean or straightforward descriptions, thereby simulating a more realistic and challenging learning environment for the VLMs. 

Our initial results show promising improvements in the model's ability to generalize attribute recognition across different datasets. We perform extensive evaluations through various datasets, assessing the impact of different types of object descriptions on model performance. Specifically,  we show a significant gap of almost 70\% in classification accuracy between testing a description in which the class name exists and testing a description which is only based on semantic object definition, without the class name. 
%
We also show that enhancing CLIP training with pure descriptive visual objects can clearly mitigate this gap. Furthermore, a modified CLIP architecture, which leverages multiple resolutions, improves the detection of fine-grained part attributes, contributing to the model's enhanced understanding of detailed visual features.


In conclusion, our research makes several contributions to exploring the limits of vision-language models in understanding descriptions of objects. 
First, we introduce a novel task, termed {\em zero-shot real object classification by description}, along with the corresponding benchmarks, to expose the current limitations of prevalent VLMs, such as recent CLIP versions, in this context. This task challenges these models to go beyond mere object recognition, focusing on intricate descriptions of parts and attributes. 
Second, we contribute to the community by releasing a dataset comprising descriptions without object names for six popular fine-grained benchmarks. This dataset can trigger research on zero-shot real classification, fostering further model development. We also propose a training method that enhances CLIP's detection of object-part attributes. This method involves curating a selection of objects from ImageNet21k and matching them with a rich array of essential part-attribute features, as generated by LLMs. This method broadens CLIP's scope of part-attribute recognition.
Third, we introduce a modified CLIP architecture that leverages multiple resolutions to enhance the model's ability to detect fine-grained part attributes. This architectural modification improves CLIP's performance in identifying details of object parts, such as shape and color, which are crucial for part attribute recognition. While the improvement in zero-shot classification tasks is modest, the enhancement in detecting fine part attributes is large, showcasing the benefits of this architectural change.


\section{Related Work}
\label{sec:related_work}

{\bf Zero-shot classification by description.} 
Recent advancements in zero-shot classification by description have significantly benefited from synergies between VLM and LLM, marking a notable change in visual classification strategies. Pioneering efforts such as those of~\cite{menon2022visual} and~\cite{Kaul2023} have established a foundational framework for using descriptions generated from LLM to enhance visual classification, establishing a new challenge for subsequent research. Although there has been notable work in both zero-shot and few-shot learning domains, including contributions from~\cite{yang2023language, zhang2023prompt,liao2023descriptor,tian2023argue}, the focus of this paper remains predominantly zero-shot classification. This stream of research has explored various innovative methodologies, such as generating customized prompts for zero-shot image classification as seen in~\cite{pratt2023does}, and learning concise descriptive attributes for visual recognition, a method further investigated by~\cite{yan2023learning}. 
A recurring theme across these works, particularly those dedicated to zero-shot learning, is the inclusion of the class name within descriptions and attributes, a practice that, while effective, raises questions about the models' reliance on explicit class identifiers. Our review of the literature reveals that to date, all known work to us on zero-shot classification by description involves descriptions and attributes to which the class name is added, underscoring a critical area for further exploration and development in the field.

{\bf Known limitations for CLIP.} 
Recent critiques of vision language models such as CLIP have pinpointed significant limitations in their processing depth, particularly highlighting their tendency to operate akin to 'bags-of-words'~\cite{yuksekgonul2022and}. This metaphor, supported by findings in the literature, underscores a superficial handling of linguistic content, where models fail to grasp complex compositional information. Specifically, the 'WaffleCLIP' study~\cite{roth2023waffling} introduced a paradigm in which LLM-generated descriptors were replaced with random words placed together with the object class name, but achieved comparable zero shot visual classification results, challenging the presumed necessity of semantic depth that is additional to class name in these models. These studies, along with the systematic breakdown of CLIP's compositional understanding failures (e.g.,~\cite{QI2023103510}), suggests that despite their versatility, VLMs may not fully comprehend the intricate interplay between textual descriptions and visual content. Our work builds upon these critiques, particularly focusing on how these models, including CLIP, might be enhanced toward a more profound and nuanced understanding of descriptions and attributes. 

{\bf Architectural changes for multi-resolution.}

Multi-scale representation has long been used as an effective technique to enhance vision tasks, \cite{lin2017featurepyramidnetworksobject,ronneberger2015unetconvolutionalnetworksbiomedical,tompson2015efficientobjectlocalizationusing}. With the advent of Vision Transformers (ViTs), various multiscale ViT architectures have emerged \cite{yang2021focalselfattentionlocalglobalinteractions,chen2021crossvitcrossattentionmultiscalevision}. More recently, \cite{shi2024needlargervisionmodels} investigated multi-scale representations as a general scaling method, achieving notable success. 
Encouraged by the findings of \cite{shi2024needlargervisionmodels}, we propose here to further fuse features from multi-resolution image processing with an additional CLIP layer, which we initially pre-train on the dataset from \cite{sharma2018conceptual} and subsequently fine-tune on our attribute-specific datasets.

\section{CLIP Description Deficiency}
\label{lebl:deficiency}
\label{sec:clip_deficiency}
\begin{table*}[]
\small
\begin{center}
\begin{scriptsize}
\begin{sc}
\begin{tabularx}{\textwidth}{X *{12}{>{\centering\arraybackslash}X}} 
\toprule
Data set & \multicolumn{3}{c}{CLIP-ViT-B-32} & \multicolumn{3}{c}{CLIP-ViT-B-16} & \multicolumn{3}{c}{CLIP-ViT-L-14} & \multicolumn{3}{c}{ConvNext} \\
\cmidrule(lr){2-4} \cmidrule(lr){5-7} \cmidrule(lr){8-10} \cmidrule(lr){11-13}
          & Only Name & With Name & No Name & Only Name & With Name & No Name & Only Name & With Name & No Name & Only Name & With Name & No Name \\
\midrule
Dogs120    & 54.0 & 58.0 & 16.9 & 57.7 & 63.9 & 18.7 & 65.6 & 72.6 & 26.5 & 58.0 & 56.9 & 17.8 \\
OxfordPets & 81.6 & 86.8 & 42.9 & 83.9 & 90.3 & 44.2 & 87.9 & 93.8 & 49.0 & 86.3 & 87.6 & 41.6 \\
Cars196    & 54.2 & 55.8 & 9.4 & 58.6 & 63.1 & 10.9 & 72.1 & 74.3 & 12.1 & 79.9 & 82.7 & 12.0 \\
CUB        & 51.5 & 52.9 & 15.1 & 55.6 & 59.0 & 17.8 & 62.6 & 64.0 & 19.8 & 68.3 & 69.3 & 16.6 \\
Flowers102 & 58.8 & 62.9 & 21.4 & 64.0 & 67.2 & 22.6 & 72.3 & 74.0 & 24.4 & 68.3 & 65.4 & 18.7 \\
Food101    & 62.1 & 68.2 & 55.7 & 7T1.4 & 76.5 & 53.4 & 78.7 & 85.4 & 65.0 & 60.4 & 58.1 & 37.1 \\
\bottomrule
\end{tabularx}
\end{sc}
\end{scriptsize}
\end{center}
\vskip -0.1in
\vspace{-1em} 
\caption{Zero-shot classification accuracy, with and without class names, Oxford prompting style.}
\label{tab:with_vs_without_names_ox}
\end{table*}

\begin{table*}[]
\begin{center}
\small
\begin{scriptsize}
\begin{sc}
\begin{tabularx}{\textwidth}{X *{12}{>{\centering\arraybackslash}X}} 
\toprule
Data set & \multicolumn{3}{c}{CLIP-ViT-B-32} & \multicolumn{3}{c}{CLIP-ViT-B-16} & \multicolumn{3}{c}{CLIP-ViT-L-14} & \multicolumn{3}{c}{ConvNext} \\
\cmidrule(lr){2-4} \cmidrule(lr){5-7} \cmidrule(lr){8-10} \cmidrule(lr){11-13}
          & Only Name & With Name & No Name & Only Name & With Name & No Name & Only Name & With Name & No Name & Only Name & With Name & No Name \\
\midrule
Dogs120    & 54.0  & 55.8 & 2.7 &  57.7& 62.3 & 4.5 & 65.6 & 70.1 & 5.5 & 58.0 & 56.1 & 6.1 \\
OxfordPets & 81.6  & 87.3 & 16.0 & 83.9 & 89.6 & 16.2 &  87.9 & 93.4 & 17.1 & 86.3 & 88.9 & 22.4 \\
Cars196    & 54.2  & 55.4 & 5.3 & 58.6 & 60.7 & 7.1 & 72.1 & 72.9 & 7.6 & 79.9 & 81.7 & 6.1 \\
CUB        & 51.5  & 51.5 & 3.7 &  55.6& 57.7 & 5.1 & 62.6 & 63.7 & 3.6 & 68.3 & 68.7 & 5.9 \\
Flowers102 & 58.8  & 64.2 & 6.2 &  64.0& 67.9 & 8.7 & 72.3 & 75.8 & 5.6 & 68.3 & 68.0 & 9.3 \\
Food101    & 62.1  & 67.5 & 25.6 & 71.4 & 75.2 & 32.3 &  78.7 & 84.7 & 40.4 & 60.4 & 58.3 & 20.4 \\
\bottomrule
\end{tabularx}
\end{sc}
\end{scriptsize}
\end{center}
\vskip -0.1in
\vspace{-1em} 
\caption{Zero-shot classification accuracy, with and without class names, Columbia prompting style.}
\label{tab:with_vs_without_names_col}
 \end{table*}

A common recent method for classification by description is to use a visual language model such as CLIP to link images with class descriptions that are generated by an LLM such as ChatGPT. This approach has been emphasized in recent studies such as \cite{menon2022visual, Kaul2023, chen2023ovarnet, roth2023waffling,pratt2023does}.
Generally, these methods incorporate the object class name within the description, potentially leading to a scenario where the class name itself heavily influences the text-to-image embedding alignment in models like CLIP. This raises a concern that the rest of the description may become secondary and the classification accuracy might be predominantly driven by the class name, now embedded within a broader textual context. This hypothesis, which we propose and explore, is supported in recent studies such as WaffleCLIP~\cite{roth2023waffling} and the analysis of CLIP as a 'Bag-of-Words'\cite{yuksekgonul2022and}, both of which examine the actual contribution of attribute-based descriptions beyond the class name in the classification by description process.

\subsection{CLIP is bad at detecting part's attributes}



To further evaluate the efficacy of the CLIP model in recognizing object and part attributes, we conducted experiments to explore its sensitivity to variations in semantic descriptions. First, using the CUB dataset (refer to Sec. \ref{sec: exp}), we generated a series of descriptions for each class in the dataset, using GPT-4 (text-only). To reduce the noise in the descriptions, we further queried GPT-4 on each attribute from the initial descriptions, ensuring a set of validated positive attributes for each class.

Alongside each primary (positive) description, five contrasting (negative) descriptions were also generated to serve as contrastive examples. For instance, if the positive description for a class was 'A bird with red wings,' the negative descriptions included variants such as 'A bird with green wings' or 'A bird with purple wings.'
Finally, to establish an upper bound for performance in this task, we evaluated GPT-4-vision, a strong multimodal large language model (MMLM) for identifying visual attributes, comparing its accuracy in recognizing positive descriptions against CLIP's performance.


Our findings, as presented in Table \ref{table:clip_part_deficiency} in columns {\em CLIP Baseline} and column {\em GPT-4-vision}, reveal limitations in the precision of the CLIP model to discern the shape, color, or size of parts of the object. The accuracy in identifying shapes, with the exception of the {\em body shape} element, resembled random chance. Color and size recognition exhibited only slight improvement over random. The average accuracy for all parts and all attributes is ~$0.33$ (single correct attribute answer was given out of six choices, that is, the random answer accuracy is $0.166$), significantly behind the CUB dataset's benchmark for zero-shot classification by CLIP, which is 63\% accuracy across 200 classes. In the next sections, we return to this part's attributes benchmark and test our fine-tuned version on the same tasks, demonstrating consistent improvement (Sec. \ref{sec:method}).
Moreover, we evaluate CLIP's effectiveness in recognizing attributes of objects and their parts at a much larger scale, via the Paco dataset (detailed in Sec. \ref{sec: exp}). 

\begin{table}[t]
\centering
\small
\begin{tabularx}{\columnwidth}{lXXXX}
\toprule
\textbf{Type} & \textbf{Element} & \textbf{CLIP Baseline} & \textbf{CLIP Finetuned} & \textbf{GPT-4-vision} \\
\midrule
Color & Back    & 0.34 & 0.39 & 0.63 \\
 & Wings   & 0.38 & 0.39 & 0.74 \\
 & Body    & 0.50 & 0.62 & 0.78 \\
 & Head    & 0.33 & 0.45 & 0.71 \\
 & Belly   & 0.35 & 0.20 & 0.73 \\
 & Average & 0.38 & 0.46 & 0.72 \\
\midrule
Shape & Beak    & 0.15 & 0.09 & 0.76 \\
 & Bill    & 0.31 & 0.53 & 0.81 \\
 & Body    & 0.41 & 0.32 & 0.87 \\
 & Head    & 0.02 & 0.33 & 0.92 \\
 & Wings   & 0.10 & 0.62 & 0.69 \\
 & Average & 0.23 & 0.33 & 0.81 \\
\midrule
Size  & Wings   & 0.17 & 0.52 & 0.63 \\
  & Neck    & 0.59 & 0.37 & 0.80 \\
  & Body    & 0.38 & 0.50 & 0.86 \\
  & Bill    & 0.40 & 0.43 & 0.57 \\
  & Tail    & 0.18 & 0.52 & 0.82 \\
  & Average & 0.38 & 0.45 & 0.74 \\
\bottomrule
\end{tabularx}
\vspace{-1em} 
\caption{Performance comparison of CLIP Baseline, CLIP Finetuned (our method), and GPT-4-vision on various CUB attribute classifications. Note that the random accuracy for this task is 0.16. It is evident that our method consistently outperforms the baseline CLIP.}
\label{table:clip_part_deficiency}
\end{table}

\subsection{Classification with and without class names}
\begin{figure}[]
\vskip 0.2in
\begin{center}
\centerline{\includegraphics[width=0.65\columnwidth]{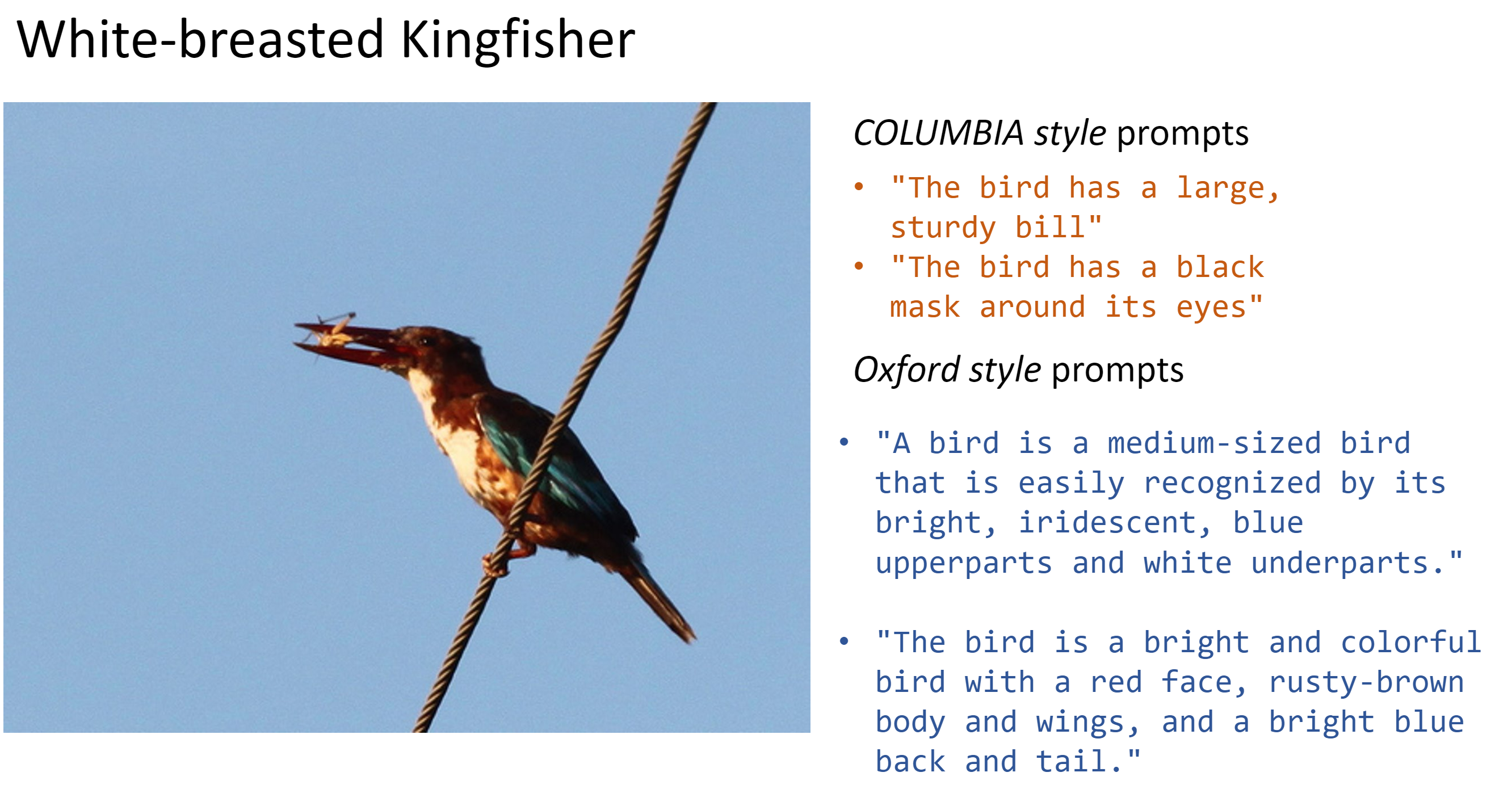}}
\caption{
{\bf Columbia and Oxford Style Descriptions Without Names.} 
 Examples highlight the difference between Columbia's concise, focused descriptions and Oxford's broader, narrative-driven approaches, both omitting object names.
 The descriptions for each example were created using one of 2 styles — the \textit{Columbia} style and the \textit{Oxford} style. Each style is a method to prompt the LLM for descriptions (usually ~8 sentences are created from each prompt style)
 }
\label{fig:examples_with_no_name_ox_and_col}
\end{center}
\vskip -0.2in
\end{figure}

In our exploration of CLIP's descriptive classification abilities, we turned our attention to the impact of including or omitting object class names from descriptions. This study was essential to understand how much the presence of class names influences the model's performance in zero-shot classification tasks. We designed experiments incorporating two distinct styles of object descriptions:
{\bf Oxford Prompting Style} (named after~\cite{Kaul2023}). This style involves generating sentences that describe objects on the basis of their general attributes and parts. The descriptions are more free-form and provide a comprehensive narrative covering various aspects of the object. However, it should be noted that generated sentences may often include repetitions. 
{\bf Columbia Prompting Style} (named after~\cite{menon2022visual}). In contrast, the Columbia style adopts a more structured approach. Its descriptions are concise, focusing specifically on a single part and attribute of the object. Unlike the Oxford style, the Columbia style's descriptions are typically non-repetitive, with each description highlighting a different aspect of the object's visual appearance.
Examples of these description styles can be found in Fig. \ref{fig:examples_with_no_name_ox_and_col}.

To rigorously test the impact of class names on classification accuracy, we applied the method described in Section \ref{lebl:method}, removing class names from descriptions in six standard fine-grained zero-shot classification datasets for Oxford and Columbia styles. Our experiments, whose results are presented in Tables \ref{tab:with_vs_without_names_ox} and \ref{tab:with_vs_without_names_col}, demonstrate a significant drop in classification efficacy when class names are excluded. This finding challenges the assumption that CLIP visually recognizes descriptive attributes and highlights the pivotal role that object names play in current zero-shot classification works.

\section{Method}
\label{lebl:method}
\label{sec:method}

This section aims to describe our approach to improving the ability of CLIP to understand the attributes of internal parts. We begin by investigating the potential benefits of training the model with a wider and more diverse range of attribute data. This approach is based on the hypothesis that diversifying CLIP training with a wide range of attribute-rich data sets could improve performance in fine-grained classification scenarios. It is important to note that the training process deliberately employs data different from the test sets; the classes used during training are different from those in the testing phase. Additionally, the attribute data used in training, while rich and varied, is generated synthetically, as we show below, and therefore may contain noise. However, our analysis shows that the effect of this noise does not seem to harm the training. 

Next, we turn to modify the CLIP architecture in order to improve its ability to detect fine object details. Our approach is inspired by \cite{shi2024needlargervisionmodels}, employing CLIP at higher resolutions and adding an attention layer to leverage multi-resolution features, and improving detection of fine details. 
In line with this methodology, a critical aspect of our training process involves the use of descriptions from which the name of the object has been deliberately omitted. By excluding the object class names, we shift the focus of CLIP's learning process towards a deeper understanding of object attributes, rather than simply associating latent image features with object classes.

\subsection{Description Creation and Name Removal}
\label{sec:data_preparation}
The name removal process was executed using a simple text processing algorithm (Algorithm 1 in Supplementary). This algorithm takes a category and its corresponding description and a default placeholder name (a super-category) as input. For example, the placeholder name for "Rhinoceros auklet" would be "bird". It begins by asking an LLM to replace any occurrence of the object name in the description with the default one. Afterwards, we initialize a list of base names derived from the category, considering various case and spacing variations. Then each base name is identified within the description and replaced with the default placeholder. This method aims to ensure that the resulting descriptions do not contain any specific class names but retain the rich descriptive attributes of each object. We have applied this approach to the two description prompting styles suggested by \cite{menon2022visual} (the {\em Oxford style}) and \cite{Kaul2023} (The {\em Columbia style}).

\subsection{Selection and Fine-Tuning on ImageNet21k}
\begin{figure}[t]
\vskip 0.2in
\begin{center}
\centerline{\includegraphics[width=0.65\columnwidth]{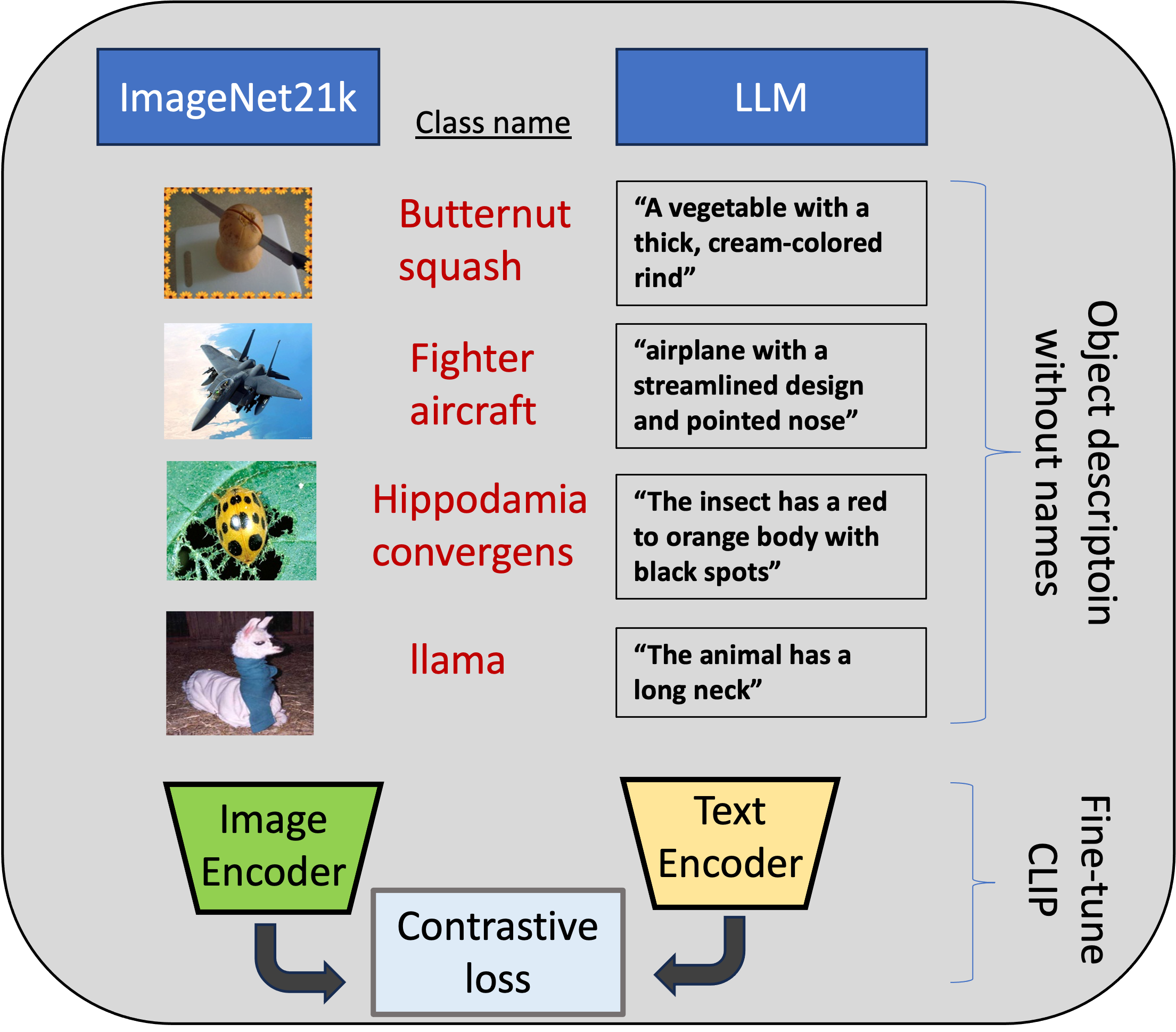}}
\caption{Real zero-shot training on ImageNet21k. This figure illustrates the process of preparing and conducting the CLIP model training for improved real zero-shot classification. The procedure begins with the selection of classes from the ImageNet21k dataset, focusing on a diverse range of object categories. For each selected class, image examples are gathered alongside the corresponding attribute descriptions generated by a Large Language Model (LLM), emphasizing the attributes of parts without including the object class names. 
}
\label{fig:imagenet21k_finetuning}
\end{center}
\vskip -0.2in
\end{figure}

To train CLIP on a comprehensive dataset of object attributes that excludes our test benchmarks, we turned to synthetic data curation. Recognizing the absence of suitable human-annotated datasets for this purpose, we chose the ImageNet21k dataset~\cite{ridnik2021imagenet} for its wide variety of object categories and subcategories, which span a wide spectrum of types. 
ImageNet21k includes numerous subcategories and provides an extensive range of attributes, especially those related to parts of objects. This diversity is important to our goal of improving attribute recognition in CLIP. In our selection process, we ensured that none of the classes appearing on our test benchmarks was included in the training set. This was a deliberate choice to prevent any direct learning of the test classes and to truly test the model's ability to generalize attribute recognition to new, unseen classes. Figure~\ref{fig:imagenet21k_finetuning} illustrates this fine-tuning process.
\begin{figure*}[t]
\vskip 0.2in
\vspace{-1em}
\begin{center}
\centerline{\includegraphics[width=1.0\textwidth]{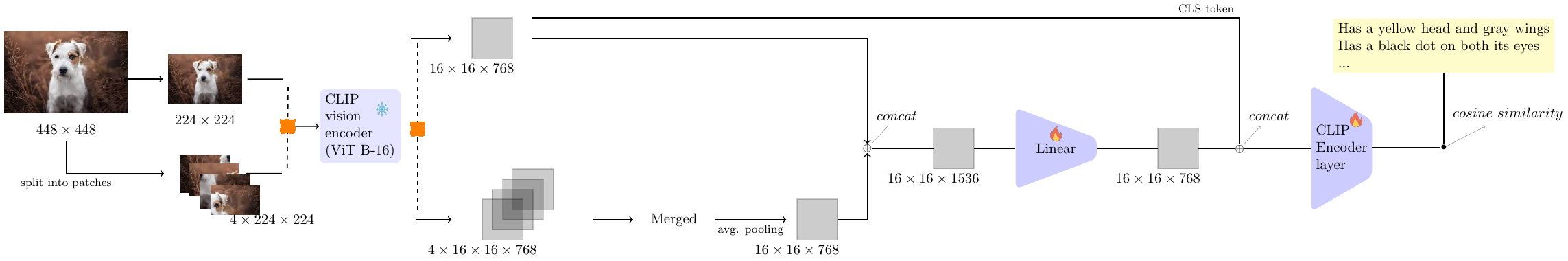}}
\caption{Our Multi-Res CLIP architecture. 
 Multiple image slices are processed via the CLIP Vision model, and multi-resolution features are aggregated using an additional CLIP Vision layer. The original CLIP model remains frozen and only the new layer is being trained.
 }
\label{fig:multires_arch}
\end{center}
\vskip -0.2in
\end{figure*}

In terms of data selection, we chose $K$ images from each category of the class, and, for each class, we generated $N$ unique sentences that describe its attributes. This resulted in $K\times N$ image-text pairs for each class. The selected classes encompassed a diverse range of animals, plants, and man-made objects (refer to Fig.~\ref{fig:comprehensive_views} A for a visual representation of the class distribution). Descriptions typically include a significant number of internal parts. The attributes covered in the descriptions ranged from various types, such as shape, color, texture, size, and other general attributes, depending on the description style used.


 \subsection{Enhancing CLIP Architecture Using Multiple Resolutions}
 

To improve CLIP’s ability to detect fine features, such as attributes of object parts, we modified the CLIP architecture to utilize multiple resolutions. This architectural enhancement, combined with the synthetic training data described above, aims to enable CLIP to perform better on tasks that require understanding of detailed attributes, ultimately enhancing its performance in real zero-shot classification scenarios. By incorporating multiple resolutions, the model can capture both coarse- and fine-grained details, providing a more comprehensive understanding of the attributes and parts of the object.
For instance, detecting the subtle curvature and color variations in a bird’s beak requires analyzing high-resolution features, which is facilitated by our multi-resolution approach.

We drew inspiration from the recent work of \cite{shi2024needlargervisionmodels} to enhance the CLIP architecture by using multiple resolutions. To describe our method, we assume a resolution of 448x448 for simplicity, although higher resolutions would function similarly. Denoting the original image by $I \in \mathbb{N}_0^{3 \times 448 \times 448}$, we first split it into 4 slices, that is,
\begin{equation*}
  I_{s_i} \in \mathbb{N}_0^{3 \times 224 \times 224}, \quad i = 1, \ldots, 4.
\end{equation*}

Denoting the number of patches in CLIP's ViT by $P$ and the visual embedding dimension by $e_d$, we input each $I_{s}$ into the CLIP vision encoder to obtain a feature map, 
\begin{equation*}
F_{s_i} = ViT(I_{s_i}) \in \mathbb{R}^{P \times e_d}
\end{equation*}

We then exclude the CLS tokens from $F_{s_i}$ and average the patch representations as follows:
\begin{equation*}
F_{AveRes}^j = \frac{1}{N} \sum_{i=1}^{N} F_{s_i}^j, \quad j=1, \ldots, P-1
\end{equation*}
where $F_{s_i}^j$ represents the $j$-th patch representation from the $i$-th slice, and $N$ is the number of slices.

To go beyond local integration and enrich the patch information with global image processing, similar to the approach in \cite{shi2024needlargervisionmodels}, we concatenate the appropriate patches. Specifically, we combine the averaged patch representations with the original patch representations as follows:
\begin{equation*}
F_j = F_{AveRes}^j \odot F_{origin}^j, \quad F_j \in \mathbb{R}^{2 \times e_d}
\end{equation*}
Where $F_{\text{origin}}^j$ is the original patch representation, and $(\odot)$ denotes concatenation.

We then use a learnable linear layer to transform $F_j$ to the original embedding dimension, 
\begin{equation*}
\Tilde{F_j} = Linear(F_j), \quad F_j \in \mathbb{R}^{e_d}
\end{equation*}

Finally, we feed the obtained feature map \(\Tilde{F_j}\) to a new, additional, and learnable ViT layer:
\begin{equation*}
\hat{F_j} = ViTLayer(\Tilde{F_j}), \quad j = 1, \ldots, P
\end{equation*}
and return the enriched CLS image token $\hat{F}_{CLS}$ as the final visual embedding. The whole process is depicted in Fig. \ref{fig:multires_arch}.

To initialize the newly introduced linear projection and attention layer, we conducted a preliminary training phase using the dataset from \cite{sharma2018conceptual}. During this phase, we keep all other components frozen and optimize only these additional layers. 
Subsequently, we fine-tune these newly trained layers, along with selected components of the original CLIP model, using the ImageNet21k dataset and the method in Sec.~\ref{sec:data_preparation}. To speed up pre-training, we initialize the added CLIP Vision Encoder layer with the same parameters as the last layer in the original CLIP Vision Encoder. 

\section{Experiments and Results}
\label{lebl:exp}
\label{sec: exp}
{\color{red}
\begin{table}[]
\begin{center}
\small
\begin{sc}
\begin{tabular}{lccc}
\toprule
Metric & CLIP-ViT-B-16  \\
\midrule
Pretrained CILP (Baseline) & 33.22  \\
Finetuned CLIP & 43.07  \\
Finetuned mluti-res CLIP & 49.84 \\
\bottomrule
\end{tabular}
\end{sc}
\end{center}
\vskip -0.1in\
\vspace{-1em} 
\caption{Avg. classification Top-1 accuracy for attribute value classification, PACO dataset.}
\label{tab:paco_experiment}
\end{table}
}
\begin{table*}[]
\begin{center}
\begin{small}
\begin{sc}
\begin{tabularx}{\textwidth}{lXXXXX}
\toprule
Data set & CLIP-ViT-B-16 & CLIP-ViT-B-16 finetuned & CLIP-ViT-B-16 Multi-Res & CLIP-ViT-L-14 & CLIP-ViT-L-14 finetuned \\
\midrule
Dogs120    & 16.9 & 20.1 & 20.4 & 26.5 & 32.8 \\
OxfordPets & 42.9 & 45.4 & 44.5 & 49.0 & 52.8 \\
Cars196    & 9.4 & 10.1 & 10.2 & 12.1 & 16.0 \\
CUB        & 15.1 & 17.7 & 18.0 & 19.8 & 24.0 \\
Flowers102 & 21.4 & 24.4 & 24.6 & 24.4 & 39.4 \\
Food101    & 55.7 & 59.0 & 58.7 & 65.0 & 69.0 \\
\bottomrule
\end{tabularx}
\end{sc}
\end{small}
\end{center}
\vskip -0.1in
\vspace{-1em} 
\caption{Classification Top-1 accuracies for real-zero shot, Oxford style.}
\label{tab:classification_results_ox}
\end{table*}

\begin{table*}[]
\begin{center}
\begin{small}
\begin{sc}
\begin{tabularx}{\textwidth}{lXXXXX}
\toprule
Data set & CLIP-ViT-B-16 & CLIP-ViT-B-16 finetuned & CLIP-ViT-B-16 Multi-Res & CLIP-ViT-L-14 & CLIP-ViT-L-14 finetuned \\
\midrule
Dogs120    & 2.7 & 11.9 & 11.8 & 5.5 & 18.1 \\
OxfordPets & 16.0 & 36.2 & 35.4 & 17.1 & 45.6 \\
Cars196    & 5.2 & 6.9 & 6.7 & 7.6 & 9.3 \\
CUB        & 3.7 & 9.3 & 9.7 & 3.6 & 16.5 \\
Flowers102 & 5.8 & 14.4 & 13.6 & 5.6 & 30.0 \\
Food101    & 30.7 & 35.7 & 35.9 & 40.4 & 46.9 \\
\bottomrule
\end{tabularx}
\end{sc}
\end{small}
\end{center}
\vskip -0.1in
\vspace{-1em} 
\caption{Classification Top-1 accuracies for real-zero shot, Columbia style.}
\label{tab:classification_results_col}
\end{table*}

This section outlines the experiments conducted to evaluate the effectiveness of our enhanced CLIP model for object and part attribute recognition. We carried out two main test series on known benchmarks:
(1) Benchmarking the original and modified CLIP models in the PACO dataset \cite{ramanathan2023paco}, a comprehensive benchmark for object part attribute recognition, to directly measure improvements at the part attribute level. 
(2) Evaluation of original and modified CLIP models in zero-shot classification against the benchmarks listed in \ref{tab:with_vs_without_names_ox}, which are standard in fine-grained classification and require visual understanding of multiple parts and their attributes. In addition, ablation studies were performed to assess the impact of various hyperparameters.

\subsection{Implementation details}
\begin{figure*} 
    \centering
    \begin{tabular}{ccc}
        \includegraphics[width=0.32\textwidth]{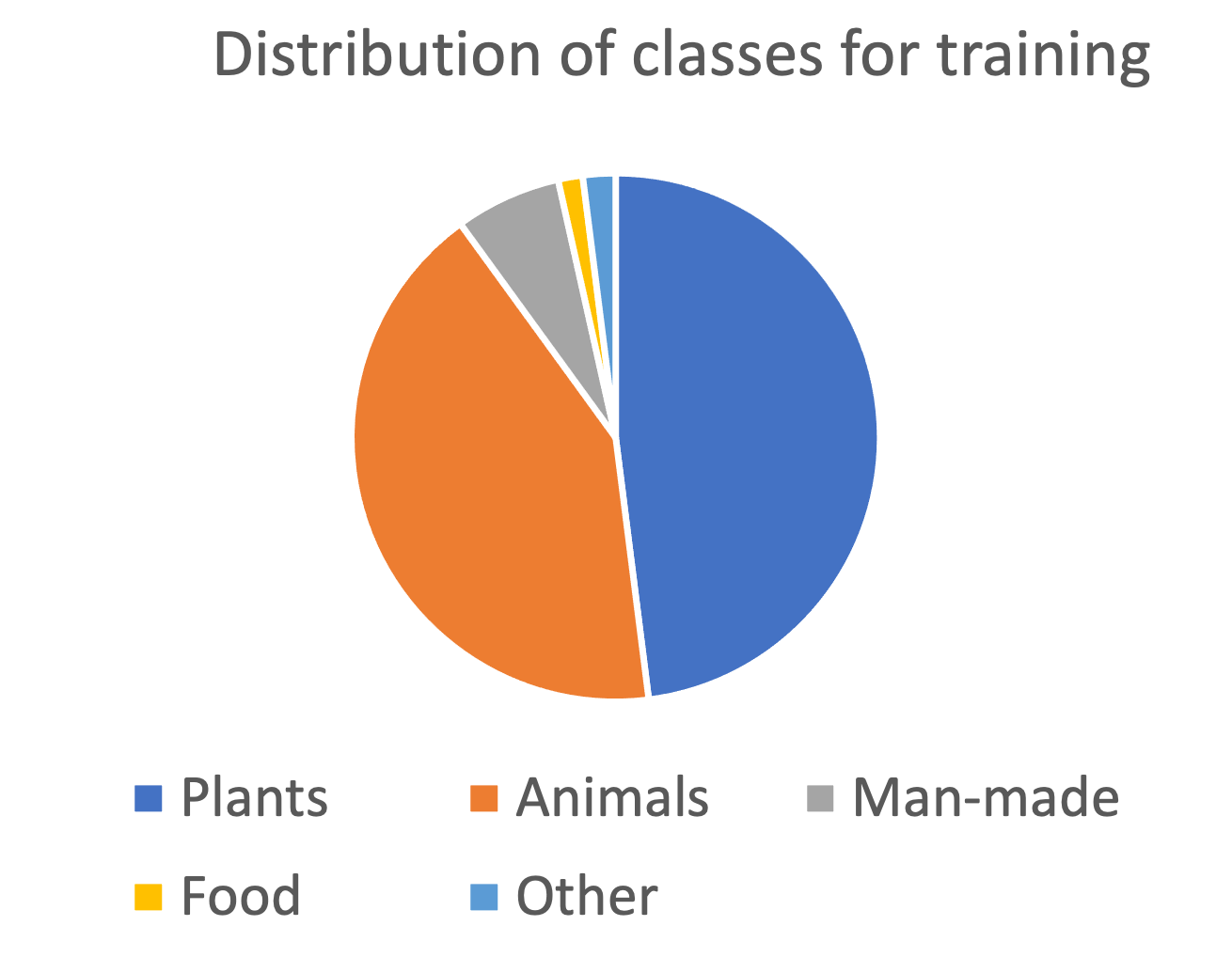} &
        \includegraphics[width=0.32\textwidth]{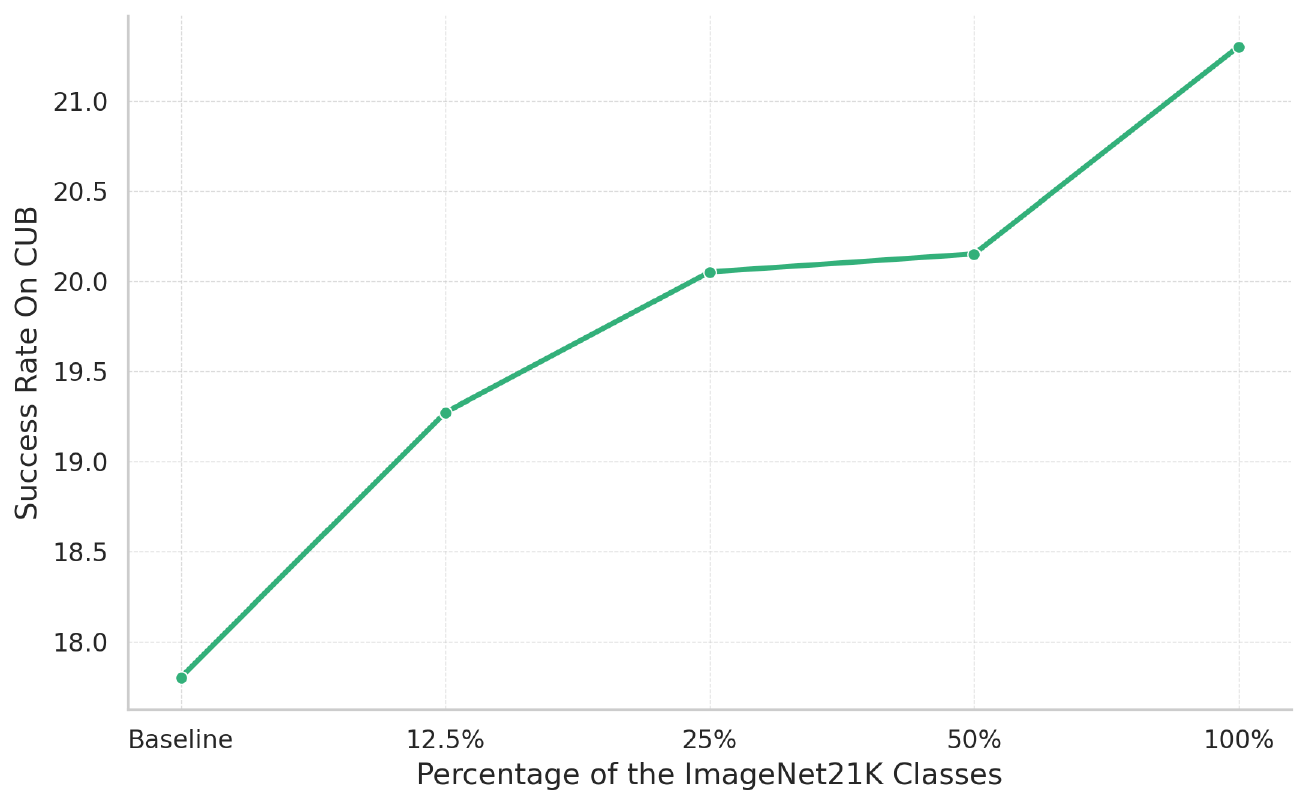} &
        \includegraphics[width=0.32\textwidth]{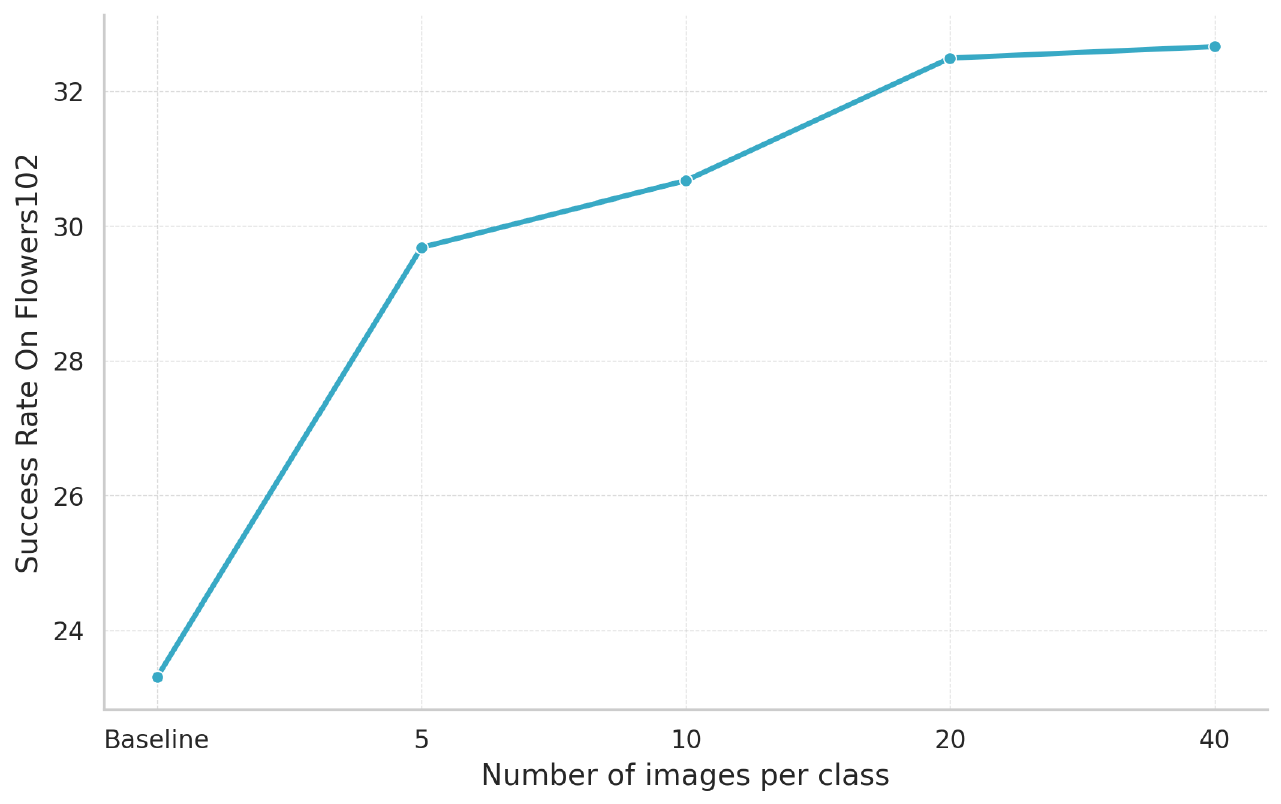} \\
        A & B & C
    \end{tabular}
    \vspace{-1em} 
    \caption{Comprehensive view of ImageNet21k training impacts. 
    {\bf (A).} The distribution of ImageNet21k classes used in training. The reason why plants and animals are dominant is because there are multiple sub-categories for these two types in ImageNet21k. These sub-categories are beneficial for our training purposes, since LLMs can provide a rich set of essential features for them.
    {\bf (B).} The impact of the number of ImageNet21k classes in the training set on zero-shot top-1 classification performance on the CUB dataset. 
    {\bf (C).} The impact of the number of images per class in the training set on zero-shot top-1 classification performance on the Flowers dataset.}
\label{fig:comprehensive_views}
\end{figure*}

\textbf{Training Data Selection.} To implement the fine-tuning procedure in Section \ref{sec:method}, we selected $K=50$ images from each category within the ImageNet21k dataset. These were matched with $N=10$ unique descriptive sentences, resulting in $N \times K$ image-text pairs for each class. Our dataset included a chosen set of $4,700$ classes, while deliberately excluding classes that were used for testing, namely categories used in fine-grained classification datasets (Sec \ref{sec:real_zero_shot_assesment}). As shown in Fig.~\ref{fig:comprehensive_views}A, the dataset is mainly composed of animals and plants, 
which constitute about 90\% of the classes. The remaining portion features man-made objects (such as types of airplane) and other general objects (such as different substances). 
An essential element in fine-tuning CLIP with this data was the selection of batches of image-text pairs, where each batch was composed of pairs from unique classes. The reason behind this was to avoid repetitions of the same caption or image within the same batch, which could damage learning from contrastive loss.


\textbf{Hyperparameter Optimization.}
We tested the learning rates of $\{1 \times 10^{-5}, 5 \times 10^{-6}, 1 \times 10^{-6}\}$ using the Adam optimizer. The batch sizes ranged from $64$ to $512$ for the ViT-Base backbone and were fixed at $108$ for the ViT-Large backbone. We fine-tuned the openai/clip-vit-base-patch32 and openai/clip-vit-large-patch14 models, averaging the results over three standard seeds: $0, 1, 2$. For all tasks, the text encoder was unfrozen during fine-tuning. For the image encoder, only the last two layers were unfrozen for fine-grained classification benchmarks, while unfreezing all layers proved to be more effective for the PACO benchmarks.

Additionally, to stabilize fine-tuning for the multi-resolution model in the classification-by-description task, we aggregate the outputs of the standard CLIP backbone and the multi-resolution encoder through a learnable weight factor. This is given by the equation 
\begin{equation}
F(I) = (1 - \alpha) \text{ViT}(I) + \alpha \hat{F}(I),
\end{equation}
where \(I\) is the image, ViT refers to the CLIP backbone encoder, and \(\hat{F}\) denotes the multi-resolution encoder. We initialize \(\alpha\) to 0.01 and set its learning rate to \(1 \times 10^{-4}\), in contrast to the learning rates of other parameters. Typically, \(\alpha\) converges to around 0.3 during training.
 

\subsection{Assessing CLIP for Object and Part Attribute Classification on PACO}

The PACO (Parts and Attributes of Common Objects) dataset \cite{ramanathan2023paco} is a comprehensive benchmark to recognize attributes of object parts. Succeeding in this benchmark is challenging, as most images contain several objects and parts, evaluating the model's precision in identifying attributes at a granular level. The dataset includes attributes of four types: {\em material}, {\em color}, {\em pattern-making}, and {\em reflectance}—each annotated at the part level. A typical annotation might be, "The rim of the bowl is black." Each part in PACO is assigned a single true value for the four attribute types, though some attribute values can be multi-valued (e.g., a color might be "blue, green"). To address these cases, we filter out objects with multiple positive attributes. Table \ref{tab:paco_experiment} shows the mean accuracy of the attribute-of-parts values, illustrating the performance comparison between the pre-trained CLIP model, its fine-tuned version, and our enhanced version of the fine-tuned modified architecture.

\subsection{Assessing CLIP for Classification by Description}
\label{sec:real_zero_shot_assesment}
To evalute the impact of improving CLIP's ability to recognize attributes on the task of classification by description, we use six fine-grained benchmark datasets popular for zero-shot description classification: CUB (Caltech-UCSD Birds-200-2011) with 200 bird species \cite{WahCUB_200_2011}, Flowers102 incorporating 102 flower categories prevalent in the UK \cite{Nilsback08}, Cars196 featuring 196 car models \cite{KrauseStarkDengFei-Fei_3DRR2013}, Food101 containing 101 food categories (each category represented by 1,000 images) \cite{bossard14}, Dogs120 comprising 120 dog breeds \cite{khosla2011novel}, and Oxford-IIIT Pet dataset, which includes 37 categories of pets with a balanced representation of cats and dogs \cite{parkhi12a}. 
Each dataset introduces specific challenges in fine-grained classification, from identifying small differences between species, breeds, or models, to discerning variations in color, shape, and texture, thereby providing a comprehensive test of the model's visual understanding capabilities. The results of this evaluation are presented in Tables~\ref{tab:classification_results_ox} and \ref{tab:classification_results_col}.


\subsection{Analysis}

The PACO results in Table \ref{tab:paco_experiment} show a consistent improvement of approximately 10\% in part attribute recognition when using the Fine-tuned CLIP model compared to the baseline CLIP. Our enhanced method, which utilizes multiple resolutions, further increases this improvement to 16\%, demonstrating the effectiveness of our approach in general attribute recognition and particularly in handling images with multiple objects.

The object classification results in Tables \ref{tab:classification_results_ox} and \ref{tab:classification_results_col} reveal significant and consistent improvements across all datasets for both the ViT-B and ViT-L versions of the fine-tuned and multiple-resolution models when using Oxford and Columbia prompting styles. 
However, the multiple resolutions approach yielded only a slight gain over the regular fine-tuned model in this case. 

For the "regular" Fine-tuned model using the Oxford style prompts, we observed performance increases of 0.7 percentage points on the Cars196 dataset, 3.3\% on Food101, and 3.2\% on the Dogs120, OxfordPets, CUB, and Flowers102 datasets. These variations can be attributed to factors such as the lower representation of objects similar to those in Cars196 in the training set, the inherent difficulty in distinguishing classes based solely on descriptions, and possibly less effective descriptions generated for Cars196 (e.g., "It is a four-door luxury vehicle with a distinct exterior and interior" for the "Acura-RL-Sedan 2012" class).

The Columbia style prompts led to even more pronounced improvements, particularly significant given the initially low top-1 percentages, with gains ranging from 1\% to impressive 25\%.

%
It should be noted that our results demonstrate consistent improvements in part and object attribute recognition across various benchmarks, sizes, and prompts. Since our fine-tuning was performed on attributes of classes not included in the test datasets, it is highlighted that the improvements observed are a result of enhanced attribute recognition rather than due to generic latent class features.

In our ablation studies, we evaluated the influence of the number of classes from ImageNet21k on the non-multiple resolution finetuned CLIP performance. Fig.~\ref{fig:comprehensive_views}B shows the results of zero-shot classification in the CUB data set with varying proportions of training classes, specifically in $12.5\%, 25\%, 50\%, 100\%$ classes used for training. The results indicate a steady increase in the success rate, correlated with the increase in the number of different classes. Furthermore, we explored the impact of increasing the number of images per class during training, employing the complete class set on the Flowers dataset. The tests were carried out with sets of $5, 10, 20, 40$ images per class, as depicted in Fig.~\ref{fig:comprehensive_views}C. The data suggests that incorporating up to 20 images per class is advantageous, with the difference in performance between 20 and 40 images being marginal.


\section{Discussion and Conclusions}
\label{lebl:discussion}

In this study, we explore the capability of CLIP in recognition of attributes of objects and their parts, as well as on the zero-shot classification task based on descriptions. Our initial investigations revealed significant limitations of CLIP in recognizing attributes of objects and their parts through carefully crafted experiments. Subsequently, we proposed a training approach and a modified CLIP architecture that improved performance on various fine-grained datasets. This result is noteworthy considering that the training data were automatically curated, which might have introduced some noise.

However, an inherent limitation of our work stems from the late-fusion architecture of CLIP. This design implies that the image encoder lacks direct awareness of the specific image regions to which the text descriptions refer, necessitating inefficient encoding of extensive data into its latent space. An intriguing avenue for future research involves exploring advances with versions of CLIP that also integrate spatial localization, such as extensions of AlphaCLIP~\cite{sun2023alpha}.




\section*{Acknowledgments}
This work was supported in part by the DARPA ECOLE program. The authors would like to thank DARPA for their support and guidance throughout this project.

{\small
\bibliographystyle{ieee_fullname}

}

\end{document}


\section{Supplementary Section}

\subsection{Algorithm 1 - Filter Name Function}

\begin{algorithm}[H]
   \caption{Filter Name Function}
   \label{alg:filter_name}
\begin{algorithmic}
   \STATE {\bfseries Input:} category $category$, description $description$, default name $default$
   \STATE {\bfseries Output:} Modified description $filtered\_description$
   \STATE
   \STATE Ask LLM to replace all occurrences of the object name in the sentence with the default name and note the output as $LLMCleanDescription$
   \STATE Initialize list of base names $baseNames$ from $category$ with case and spacing variations
   \FOR{each name $name$ in $baseNames$}
       \STATE Replace occurrences of $name$ in $LLMCleanDescription$ with $default$
   \ENDFOR
   
   Return $filtered\_description$
\end{algorithmic}
\end{algorithm}

\subsection{PACO experiment}
PACO (Parts and Attributes of Common Objects) dataset \cite{ramanathan2023paco} is an extensive benchmark for object parts classification, detection, and segmentation. PACO contains attributes of four "types" ($material$, $color$, $pattern-making$, $reflectance$) which are multi-valued and at the \textit{part level}. For instance, a single annotation out of the dataset may be considered as "The \texttt{part} of \texttt{object} has \texttt{attribute\_value} \texttt{attribute\_type}", For instance "The rim of the bowl has black color"

The dataset is structured such that for each part there is a single true value for the 4 mentioned attribute types. However, some of the \texttt{attribute\_values} are multi-valued, and in particular are composed of several other values. For instance, a "color" may be "blue, green" for certain parts. That causes a problem when creating an attribute classification, since the values "blue", "green" are also contained within the possible values of "color". In order to decompose such cases, we have considered \textit{both} "blue" and "green" as positive attribute values for the respective part. Furthermore, some of the parts are edges of other parts of objects, which are barely noticeable in the image (only a few pixels wide), which we had removed from the test set.

In order to produce the test, we have attached the sentences of the above pattern for each part and attribute of the PACO test set. Subsequently, we evaluated our model's understanding of the attributes for each part by assessing similarity scores for every image and part. Therefore, the predicted value for \texttt{attribute\_value} of a certain image part and attribute type would be determined by the maximizing clip similarity among all possible values (of that \texttt{attribute\_type} in the entire dataset). To make a fair comparison, we had predicted for each combination of (\texttt{image}, \texttt{part}, \texttt{attribute\_type}) individually. In case an example has $k$ positive attribute values, we take the top $k$ values maximizing the clip similarity. We had calculated the avg. accuracy for attribute-of-parts values for the pre-trained CLIP and our fine-tuned.